\documentclass[11pt]{article}
\usepackage{paclic35}
\usepackage{latexsym}
\usepackage{amsmath}
\usepackage{multirow}
\usepackage{url}
\usepackage{float}
\usepackage{pgfplots}
\usepackage{footnote}
\usepackage[flushleft]{threeparttable}
\usepackage[utf8]{inputenc}
\usepackage[T5]{fontenc}
\usepackage{longtable}
\usepackage{supertabular}
\usepackage{makecell}
\usepackage{multicol}
\usepackage{graphicx}
\usepackage{array}
\usepackage{lipsum}

\begin{document}
\title{Monolingual versus Multilingual BERTology for Vietnamese Extractive Multi-Document Summarization}


  
  
\author{Huy Quoc To$^{1,2,3}$, \textbf{Kiet Van Nguyen}$^{1,2,3}$,\textbf{Ngan Luu-Thuy Nguyen}$^{1,2,3}$,\textbf{Anh Gia-Tuan Nguyen}$^{1,2,3}$\\
$^{1}$University of Information Technology, Ho Chi Minh City, Vietnam\\
$^{2}$Vietnam National University, Ho Chi Minh City, Vietnam \\
$^{3}$\{huytq,kietnv,ngannlt,anhngt\}@uit.edu.vn}

\maketitle
\begin{abstract}
Recent researches have demonstrated that BERT shows potential in a wide range of natural language processing tasks. It is adopted as an encoder for many state-of-the-art automatic summarizing systems, which achieve excellent performance. However, so far, there is not much work done for Vietnamese. In this paper, we showcase how BERT can be implemented for extractive text summarization in Vietnamese on multi-document. We introduce a novel comparison between different multilingual and monolingual BERT models. The experiment results indicate that monolingual models produce promising results compared to other multilingual models and previous text summarizing models for Vietnamese.
\end{abstract}

\section{Introduction}

Automatic text summarization is one of the challenging tasks of natural language processing (NLP). This task requires the machine to generate a piece of text which is a shorter version of one or many input documents. The output of the automatic text summarization tool has to be condensed into information while preserving the primary knowledge of the given documents. The applications for this task are widely implemented. A simple example is an automatic summarizing tool that can generate abstracts for scientific research papers. This tool can save much time for researchers and also readers. Despite the fact that the abstract is just a summary of a research paper, it still costs plenty of time for researchers to read through their work and rewrite it into a concise paragraph. 
\par 
Though the text summarization in English has obtained substantial achievements, there has not been much work done in Vietnamese. In recent years, the study of Vietnamese text summarizing algorithms has been mainly on features ranking \cite{Ung2015,nguyen2016} and graph-based systems \cite{nguyen2012tsgvi}. Meanwhile, the pre-trained Bidirectional Encoder Representation from Transformers (BERT) model \cite{devlin-etal-2019-bert} has shown its significant advantages in many NLP tasks in Vietnamese such as text classifications \cite{nguyen2021constructive,nguyen2021vietnamese} and machine reading comprehension \cite{luu2021mrc}. There are some empirical studies on deep learning for this task on Vietnamese datasets, which show outstanding results compared to traditional algorithms \cite{nguyen2018}. However, there is still a lack of academic work for text summarization in Vietnamese using BERT. In this paper, we aim to examine the influence of several BERT models on extractive text summarization in Vietnamese. We conduct experiments on both monolingual and multilingual BERT models to compare their performance on this task. 

Our main contributions on this paper are:
\begin{itemize}
  \item We experiment with new implementations and compare the efficiency of multiple multilingual and monolingual models on generating a summary. As a result, the viBERT4news model achieves the highest average ROUGE-1 score with 77.44\%.
  \item We also perform comparison and analyses with previous automatic summarizing systems specialized for Vietnamese text to prove the effectiveness of our proposals.
\end{itemize}

The remainder of this paper is structured as follows. In section 2, we perform a literature review on prior studies. Section 3 further describes the datasets that are used to evaluate the models. Next, we showcase our approaches in section 4 and the details of our experiments in section 5. The results are then analysed in section 6. Lastly, we conclude our work in section 7.     

\section{Previous Work}
The techniques for test summarization are divided into two categories: extraction and abstraction. An extractive summarization is a combination of sentences extracted from the original text. These sentences are calculated to carry the actual content of the document. In contrast, abstractive summarization is a technique that examines the document and produces a new text. In other words, this method creates novel sentences based on the most critical information of the originals. Even though people usually use abstractive ways to create a summary, extractive summarization methods have more attention on recent researches \cite{Allahyari2017}. The performances of extractive summarizing systems are often better than the abstractive summarizing systems \cite{erkan04}.
\par
In addition, there are two approaches for such tasks: supervised and unsupervised learning algorithms. In supervised approaches, a classifier is trained using annotated data. This classifier determines which sentences are included in the summary and which are not. In the training phase of supervised methods, the features extracted from training data are fed to the models. Then in validating and testing phases, the trained models can predict the results using unseen data \cite{cao2015,ren2017}. This approach creates opportunities for researchers to modify the models based on the experimental results in validating phases. However, the prerequisite of this approach is the predefined annotated features using for training. It remains an unresolved challenge for such a low-resource language as Vietnamese. Therefore, the prior works for text summarization in Vietnamese have been focused on the unsupervised algorithm. The main advantage of this method is that the annotated data is not compulsory. \cite{nguyen2017,banerjee2015}. 
\par
The work on unsupervised learning for extractive summarization has a long history. In most of the researches, the main algorithm applying in unsupervised learning is obtaining sentences with top-ranking scores. This method computes the score for each sentence of an input document based on several features such as sentence length \cite{10.5555/2886521.2886620}, TF-IDF \cite{erkan04}, graph-based \cite{belwal2020new}, sentence position \cite{10.1145/3077136.3080792}. 
The proposals for text summarization in Vietnamese also focus on sentence ranking. For instance, Dinh and Nguyen \cite{dinh-and-nguyen} showcased how to represent sentences as nodes in a graph. The authors employed the PageRank algorithm to evaluate the importance of the sentences, and it resulted in significant performance on both long and short documents. 

\par
On the other hand, Liu and Lapata \shortcite{liu-lapata-2019-text} proposed a novel approach for both abstractive and extractive summarization utilizing BERT. The experiments were performed on their popular datasets which are CNN/Daily Mail, NYT and XSum. The authors applied BERT as an encoder at the document level in their framework. Their report demonstrated that BERT achieved state-of-the-art results compared to both other automatic summarizing tools and human-based evaluation. Xu et al. \shortcite{xu-etal-2020-discourse} further improved the baseline BERT model in order to remove repeated and uninformative phrases in the final summary. Their research introduced a discourse-aware BERT model (DISCOBERT) that focused on discourse units instead of sentences. 
\par
The above approaches have not addressed the efficiency of BERT on Vietnamese text though there is a tremendous amount of work investigated on English. It raises a big question on how effective BERT can perform on summarizing as well as whether monolingual BERT achieves better results than multilingual BERT in Vietnamese documents, especially, in multiple documents. In this paper, we aim to conduct an empirical study on multiple unsupervised BERT-based models that combine BERT as encoders and K-means clustering as a sentence ranking algorithm. Our approach is further described in the next section.

\section{Multilingual BERT models}
 
\subsection{mBERT}
The multilingual BERT (mBERT) model is basically a BERT model trained on pre-trained on 104 languages data from Wikipedia and BookCorpus. The main architecture of the multilingual BERT model is based on the Transformer architecture. In the pre-trained phase, this model was trained on a total of 16GB of unlabelled text data so that it covers a large amount of vocabulary in many languages. In our approach, we use individual sentences as the inputs for this mBERT model. We implement two mBERT models in the experiments which are {\tt bert-base-multilingual-cased} and {\tt bert-base-multilingual-uncased}. According to Devlin et al. \shortcite{devlin-etal-2019-bert}, these two models treat the case sensitivity of words differently. For example, the words \emph{English} and \emph{english} have no distinction in the {\tt bert-base-multilingual-uncased} model, while in the cased model, these are two different words.
\subsection{XLM-RoBERTa}
The XLM-RoBERTa was first presented by Conneau et al. \shortcite{conneau-etal-2020-unsupervised} which was pre-trained on text data from 100 languages at a massive scale. This model achieves state-of-the-art performance for cross-lingual NLP tasks. The results of this model even outperform members on classification tasks by 23\% in accuracy score. XLM-RoBERTa is also specifically built for low-resource languages. Therefore, we believe that this model is an efficient choice for summarizing Vietnamese text. In the experiment, we employ two models which are {\tt xlm-roberta-base} (12 layers, 768 hidden dimensions, 12 self-attention heads, 270M parameters) and {\tt xlm-roberta-large} (24 layers, 1024 hidden dimensions, 16 self-attention heads, 550M parameters). 

\subsection{DistilBERT}
DistilBERT model is technically a cut-down version of BERT model in order to reduce the inference time but maintain the performance. This model was proposed by Sanh et al. \shortcite{Sanh2019DistilBERTAD} in 2019, and it has demonstrated some potential. The author showcased that DistilBERT obtains up to 97\% performance of BERT models. However, it noticeably runs 60\% faster with 40\% fewer parameters. We use {\tt distilbert-base-multilingual-cased} (6 layers, 768 dimension and 12 heads, 134M parameters), which is a distilled version of mBERT in our comparison.

\section{Monolingual BERT models}
\subsection{PhoBERT}
In terms of monolingual BERT models specifically implemented for Vietnamese, PhoBERT is well-known among all. PhoBERT is developed by Nguyen and Nguyen \shortcite{nguyen2020phobert} with two versions, {\tt PhoBERT-base} and {\tt PhoBERT-large} based on the architectures of {\tt BERT-large} and {\tt BERT-large} respectively. The models were pre-trained on approximately 20GB of textual data from the Vietnamese news corpus\footnote{https://github.com/binhvq/news-corpus} and the Vietnamese Wikipedia corpus. PhoBERT has the same training approach as RoBERTa \cite{liu2019roberta} which is implementing Dynamic Masking to create masked tokens whenever a sentence is fed in the model.
Since these models are trained on the Vietnamese dataset, they gain outstanding results on several NLP tasks in Vietnamese such as machine reading comprehension \cite{luu2021mrc} and text classification \cite{nguyen2021constructive}.
\subsection{Bert4News}
Bert4News model was developed by researchers from the Hanoi University of Science and Technology. It was pre-trained on 20GB of Vietnamese news data with a vocabulary size of 62,000 words. The tokenization method used in Bert4news is based on syllables which are proven to be an effective way to define the basic unit in Vietnamese text. This model also achieved high accuracy on text segmentation and named entity recognition in Vietnamese. We download {\tt vibert4news-base-cased} to implement this model in our settings.

\section{Experiments}

\subsection{K-means clustering}
To rank and select the sentences for the final summary, we apply the K-means clustering algorithm. Since we investigate the text summarization on multiple documents, for each topic, there are at least two centroids and many embedded sentences around them as the result of the BERT embedding layer. K-means algorithm will start to select the most important sentences by choosing the embedded sentences that are closest to the centroids of each document. We utilize the Sci-kit learn library\footnote{https://scikit-learn.org} to implement this algorithm in our experiments. 
\subsection{Dataset}
The datasets using in our paper are VietnameseMDS\footnote{https://github.com/lupanh/VietnameseMDS}. To the best of our knowledge, this is the only available dataset for evaluating extractive text summarization systems in Vietnamese. The dataset contains 200 clusters representing Vietnamese news topics collected from Baomoi\footnote{https://baomoi.com} - a digital newspaper in Vietnamese. In each cluster, there are normally 2 to 5 documents that are related to one topic. Additionally, the authors provide segmented data in both sentence and word levels in a separate file. Each cluster also goes along with two gold references which are made by humans. These references are later used to evaluate the efficiency of machine summarization. Table \ref{table-overview} shows the overview information of the dataset.

\begin{table}[H]
\begin{center}
    \begin{tabular}{|c|r|}
    \hline
    \textbf{Dataset}    & \multicolumn{1}{l|}{VietnameseMDS} \\ \hline
    \textbf{Clusters}   & 200                                \\ \hline
    \textbf{Documents}  & 600                                \\ \hline
    \textbf{Sentences}  & 9,802                              \\ \hline
    \textbf{References} & 400                                \\ \hline
    \end{tabular}
\end{center}
\caption{Overview of the VietnamesMDS dataset.}
\label{table-overview}
\end{table}

\par
The length of gold references in this dataset is not identical. The average sentence length in the dataset is 50 words. According to Ung et al. \cite{Ung2015}, the length of reference can be categorized into six groups. Each group contains references that have the same sentence length in words. However, we do not apply the same division since we approach this task by identifying the best average number of sentences for the summary of our models.

\subsection{ROUGE score}
In our experiment, we use F-measure of ROUGE score to evaluate the model performance. ROUGE score is a type of evaluation metrics that is mainly used for automatic summarization system. These metrics conduct comparison between systematically produced summary and human-produced summary. In particularly, in our research, we apply ROUGE-1 AND ROUGE-2 score which are based on calculating the overlap of N-grams between the two summaries. The ROUGE-1 means the overlap of unigram and ROUGE-2 refers to the overlap of bigrams. 

\subsection{Experimental Settings}

\begin{figure}[H]

    \begin{minipage}[t]{.45\textwidth}
        \includegraphics[scale=0.48]{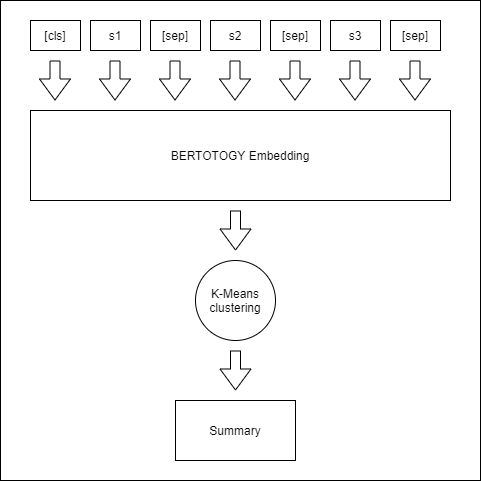}
        \centering
    \end{minipage}
    \caption{Each sentence (s1,s2,s3) is encoded by BERT models then ranking and selecting for final summary. The length of the summary can be determined by the K-means algorithm.} 
    \label{fig:bertology}
\end{figure}

To perform the comparison, we produce an architecture described in Figure \ref{fig:bertology} that comprises two main components: BERT models for embedding and K-means clustering for ranking and sentence selection for summarization. We make use of the tokenized text provided in the dataset to generate the summaries. First of all, we extract the content of the documents and store them into a list of sentences. Since our objective is to create multi-document summaries, we concatenate all the documents in one cluster into one paragraph. Then the paragraph is tokenized into sentences and passed into BERT models for embedding. All the BERT models are set to base configuration. BERT models implemented in this paper are taken from PyTorch-pre-trained-BERT libraries, which are provided on {\tt huggingface}. We continue with clustering the sentence embedding producing by BERT models. At this stage, we set the parameter {\tt K = 4} since we have found that it is the optimal number of clusters. Parameter {\tt K} represents the desired number of sentences for output.
\par

\begin{table*}[t]
\centering
\begin{tabular}{|c|l|r|r|}
\hline
                                                                                       & \textbf{Models}               & \multicolumn{1}{l|}{\textbf{ROUGE-1}} & \multicolumn{1}{l|}{\textbf{ROUGE-2}} \\ \hline
\multirow{5}{*}{\textbf{\begin{tabular}[c]{@{}c@{}}Multilingual \\ BERT\end{tabular}}} & Multilingual-BERT-uncased      & 77.34                                 &     51.68                                  \\ \cline{2-4} 
                                                                                       & Multilingual-BERT-cased        & 77.26                                 &      50.52                                 \\ \cline{2-4} 
                                                                                       & XLM-R-base                    & 77.40                                 &       50.68                                \\ \cline{2-4} 
                                                                                       & XLM-R-large                   & 77.40                                 &      51.19                                 \\ \cline{2-4} 
                                                                                       & DistilBERT-multilingual-cased & 77.42                                 &      50.76                                 \\ \hline
\multirow{3}{*}{\textbf{\begin{tabular}[c]{@{}c@{}}Monolingual\\ BERT\end{tabular}}}   & PhoBERT-base                  & 77.07                                 &       50.95                                \\ \cline{2-4} 
                                                                                       & PhoBERT-large                 & 77.42                                 &        50.89                               \\ \cline{2-4} 
                                                                                       & viBERT4news                     & \textbf{{77.44}}                              &    \textbf{52.01}                                  \\ \hline
\end{tabular}
\caption{Comparison with F-score on BERT models.}
\label{table-multi-mono-bert}
\end{table*}

\section{Results}
We report our comparison on various multilingual and monolingual BERT models in Table \ref{table-multi-mono-bert}. As there are two gold references in each cluster, we perform ROUGE-score evaluation on both references and retrieve the best F-score for each topic. From there, for each model, we are able to calculate the best average score using the best results.
\par
It illustrates the fact that the viBERT4news outperforms other models with \textbf{77.44\%} on ROUGE-1 score and \textbf{52.01\%} on ROUGE-2 score. We can conclude that this monolingual BERT model has better performance on extractive summarization tasks. However, it is not too far from the best results from multilingual BERT models. Multilingual distilBERT achieves 77.42\% on ROUGE-1 score, which is the second-best result among others. It also clearly confirms that not all monolingual models have higher results than multilingual BERT models. 

\par
We have noticed that viBERT4news was trained on 20GB of Vietnamese news data. This is concrete evidence for the reason why it performs effectively on VietnameseMDS since this dataset contains data from an online news website. 

\par
We also create a comparison between best performance BERT models with other models implementing for Vietnamese text summarization, namely LSA, LexRank, KL-divergence for unsupervised methods, as well as CNN and LSTM for deep learning methods \cite{nguyen2018}. 

Table \ref{table-bert-vs-others} shows that our model performance is promising. The gap between traditional unsupervised methods and deep learning methods with our model is substantial. In the ROUGE-1 score, the highest achievement in the unsupervised method is 60.2\%, and deep learning is 52.8\%. The results of our model are 25\% higher compared to those.

\begin{table}[H]
\centering
\label{tab:my-table}
\begin{tabular}{|l|r|r|}
\hline
\textbf{Model} & \multicolumn{1}{l|}{\textbf{ROUGE-1}} & \multicolumn{1}{l|}{\textbf{ROUGE-2}} \\ \hline
LSA            & 49.2                                  & 39.2                                  \\ \hline
LexRank        & 48.2                                  & 39.2                                  \\ \hline
KL             & 60.2                                  & 40.4                                  \\ \hline
CNN            & 52.8                                  & 40.0                                  \\ \hline
LSTM           & 52.5                                  & 39.6                                  \\ \hline
DistillBERT    & 77.4                                  & 50.8                                  \\ \hline
viBERT4news    & \textbf{77.4}                         & \textbf{52.0}                         \\ \hline
\end{tabular}
\caption{Comparison model performance of BERT and other models in F-scores.}
\label{table-bert-vs-others}
\end{table}

\par
Table \ref{table-bert-vs-tools} illustrates the results of our model comparing with two well-known systems for Vietnamese text summarization which are TSGVi \cite{tuanh2010} and CFVi \cite{Ung2015}. It is proved that BERT models yield the best performance among other models and systems. Even though the performance of other Vietnamese summarization tools is relatively high, The evaluation results of the CFVi-2 tool are 76.38\% for ROUGE-1 and 49.43\% for ROUGE-2, which are not far from BERT results.

\begin{table}[H]
\centering
\label{tab:my-table}
\begin{tabular}{|l|r|r|}
\hline
\textbf{Model} & \multicolumn{1}{l|}{\textbf{ROUGE-1}} & \multicolumn{1}{l|}{\textbf{ROUGE-2}} \\ \hline
TSGVi          & 74.14                                 & 42.34                                 \\ \hline
CFVi-1         & 75.16                                 & 45.29                                 \\ \hline
CFVi-2         & 76.38                                 & 49.43                                 \\ \hline
DistillBERT    & 77.42                                 & 50.76                                 \\ \hline
viBERT4news    & \textbf{77.44}                        & \textbf{52.01}                        \\ \hline
\end{tabular}
\caption{Comparison with previous Vietnamese text summarization systems in F-scores.}
\label{table-bert-vs-tools}
\end{table}

Due to the fact that viBERT4news model was majorly pre-trained on news data, its results outperform other models including both multilingual and monolingual models. The key factor is that the dataset and gold references used in our experiments, as mentioned, are extracted from online news. The vocabulary range used in the dataset, gold references and pre-trained viBERT4news is matched to some extend. 

\section{Conclusion and Future Work}
The paper has described a comparison between multilingual and monolingual BERT models using for Vietnamese extractive multi-document summarization. We also compare the best performance models of each type of BERT with other extractive summarizing systems and models. The results indicate that BERT models always produce a better summary, although they do not excel comparing to Vietnamese graph-based models. However, the efficiency of this approach is undeniable. 
\par
In future work, we will combine other techniques for data processing such as keyword extraction. We will also improve the models so that the output summary is closer to the human summary. After that, we will conduct experiments on other datasets in other languages to further approve our approach.

\bibliographystyle{acl}
\bibliography{references}

\begin{thebibliography}{}

\bibitem[\protect\citename{Allahyari \bgroup et al.\egroup
  }2017]{Allahyari2017}
Mehdi Allahyari, Seyedamin Pouriyeh, Mehdi Assefi, Saeid Safaei, Elizabeth~D.
  Trippe, Juan~B. Gutierrez, and Krys Kochut.
\newblock 2017.
\newblock Text summarization techniques: A brief survey.
\newblock {\em International Journal of Advanced Computer Science and
  Applications}, 8(10).

\bibitem[\protect\citename{Banerjee \bgroup et al.\egroup }2015]{banerjee2015}
Siddhartha Banerjee, Prasenjit Mitra, and Kazunari Sugiyama.
\newblock 2015.
\newblock Multi-document abstractive summarization using ilp based
  multi-sentence compression.
\newblock In {\em Proceedings of the 24th International Conference on
  Artificial Intelligence}, IJCAI'15, page 1208–1214. AAAI Press.

\bibitem[\protect\citename{Belwal \bgroup et al.\egroup }2020]{belwal2020new}
Ramesh~Chandra Belwal, Sawan Rai, and Atul Gupta.
\newblock 2020.
\newblock A new graph-based extractive text summarization using keywords or
  topic modeling.
\newblock {\em Journal of Ambient Intelligence and Humanized Computing}, pages
  1--16.

\bibitem[\protect\citename{Cao \bgroup et al.\egroup }2015a]{cao2015}
Ziqiang Cao, Furu Wei, Li~Dong, Sujian Li, and Ming Zhou.
\newblock 2015a.
\newblock Ranking with recursive neural networks and its application to
  multi-document summarization.
\newblock In {\em Proceedings of the Twenty-Ninth AAAI Conference on Artificial
  Intelligence}, AAAI'15, page 2153–2159. AAAI Press.

\bibitem[\protect\citename{Cao \bgroup et al.\egroup
  }2015b]{10.5555/2886521.2886620}
Ziqiang Cao, Furu Wei, Li~Dong, Sujian Li, and Ming Zhou.
\newblock 2015b.
\newblock Ranking with recursive neural networks and its application to
  multi-document summarization.
\newblock In {\em Proceedings of the Twenty-Ninth AAAI Conference on Artificial
  Intelligence}, AAAI'15, page 2153–2159. AAAI Press.

\bibitem[\protect\citename{Conneau \bgroup et al.\egroup
  }2020]{conneau-etal-2020-unsupervised}
Alexis Conneau, Kartikay Khandelwal, Naman Goyal, Vishrav Chaudhary, Guillaume
  Wenzek, Francisco Guzm{\'a}n, Edouard Grave, Myle Ott, Luke Zettlemoyer, and
  Veselin Stoyanov.
\newblock 2020.
\newblock Unsupervised cross-lingual representation learning at scale.
\newblock In {\em Proceedings of the 58th Annual Meeting of the Association for
  Computational Linguistics}, pages 8440--8451, Online, July. Association for
  Computational Linguistics.

\bibitem[\protect\citename{Devlin \bgroup et al.\egroup
  }2019]{devlin-etal-2019-bert}
Jacob Devlin, Ming-Wei Chang, Kenton Lee, and Kristina Toutanova.
\newblock 2019.
\newblock {BERT}: Pre-training of deep bidirectional transformers for language
  understanding.
\newblock In {\em Proceedings of the 2019 Conference of the North {A}merican
  Chapter of the Association for Computational Linguistics: Human Language
  Technologies, Volume 1 (Long and Short Papers)}, pages 4171--4186,
  Minneapolis, Minnesota, June. Association for Computational Linguistics.

\bibitem[\protect\citename{Erkan and Radev}2004]{erkan04}
G\"{u}nes Erkan and Dragomir~R. Radev.
\newblock 2004.
\newblock Lexrank: Graph-based lexical centrality as salience in text
  summarization.
\newblock {\em J. Artif. Int. Res.}, 22(1):457–479, December.

\bibitem[\protect\citename{Liu and Lapata}2019]{liu-lapata-2019-text}
Yang Liu and Mirella Lapata.
\newblock 2019.
\newblock Text summarization with pretrained encoders.
\newblock In {\em Proceedings of the 2019 Conference on Empirical Methods in
  Natural Language Processing and the 9th International Joint Conference on
  Natural Language Processing (EMNLP-IJCNLP)}, pages 3730--3740, Hong Kong,
  China, November. Association for Computational Linguistics.

\bibitem[\protect\citename{Liu \bgroup et al.\egroup }2019]{liu2019roberta}
Yinhan Liu, Myle Ott, Naman Goyal, Jingfei Du, Mandar Joshi, Danqi Chen, Omer
  Levy, Mike Lewis, Luke Zettlemoyer, and Veselin Stoyanov.
\newblock 2019.
\newblock Roberta: A robustly optimized bert pretraining approach.

\bibitem[\protect\citename{Luu \bgroup et al.\egroup }2021]{luu2021mrc}
Son~T. Luu, Kiet Van~Nguyen, Anh Gia-Tuan Nguyen, and Ngan Luu-Thuy~Nguyen.
\newblock 2021.
\newblock An experimental study of deep neural network models for vietnamese
  multiple-choice reading comprehension.
\newblock In {\em 2020 IEEE Eighth International Conference on Communications
  and Electronics (ICCE)}, pages 282--287.

\bibitem[\protect\citename{Nguyen and Nguyen}2020]{nguyen2020phobert}
Dat~Quoc Nguyen and Anh~Tuan Nguyen.
\newblock 2020.
\newblock Phobert: Pre-trained language models for vietnamese.
\newblock {\em arXiv preprint arXiv:2003.00744}.

\bibitem[\protect\citename{Nguyen \bgroup et al.\egroup }2016]{nguyen2016}
Hy~Nguyen, Tung Le, Viet-Thang Luong, Minh-Quoc Nghiem, and Dien Dinh.
\newblock 2016.
\newblock The combination of similarity measures for extractive summarization.
\newblock In {\em Proceedings of the Seventh Symposium on Information and
  Communication Technology}, SoICT '16, page 66–72, New York, NY, USA.
  Association for Computing Machinery.

\bibitem[\protect\citename{Nguyen \bgroup et al.\egroup }2017]{nguyen2017}
Minh-Tien Nguyen, Tran~Viet Cuong, Nguyen~Xuan Hoai, and Minh-Le Nguyen.
\newblock 2017.
\newblock Utilizing user posts to enrich web document summarization with matrix
  co-factorization.
\newblock In {\em Proceedings of the Eighth International Symposium on
  Information and Communication Technology}, SoICT 2017, page 70–77, New
  York, NY, USA. Association for Computing Machinery.

\bibitem[\protect\citename{Nguyen \bgroup et al.\egroup }2018]{nguyen2018}
Minh-Tien Nguyen, Hoang-Diep Nguyen, Thi-Hai-Nang Nguyen, and Van-Hau Nguyen.
\newblock 2018.
\newblock Towards state-of-the-art baselines for vietnamese multi-document
  summarization.
\newblock In {\em 2018 10th International Conference on Knowledge and Systems
  Engineering (KSE)}, pages 85--90.

\bibitem[\protect\citename{Nguyen \bgroup et al.\egroup
  }2021a]{nguyen2021constructive}
Luan~Thanh Nguyen, Kiet Van~Nguyen, and Ngan Luu-Thuy Nguyen.
\newblock 2021a.
\newblock Constructive and toxic speech detection for open-domain social media
  comments in vietnamese.
\newblock {\em arXiv preprint arXiv:2103.10069}.

\bibitem[\protect\citename{Nguyen \bgroup et al.\egroup
  }2021b]{nguyen2021vietnamese}
Nhung Thi-Hong Nguyen, Phuong Ha-Dieu Phan, Luan~Thanh Nguyen, Kiet Van~Nguyen,
  and Ngan Luu-Thuy Nguyen.
\newblock 2021b.
\newblock Vietnamese open-domain complaint detection in e-commerce websites.
\newblock {\em arXiv preprint arXiv:2104.11969}.

\bibitem[\protect\citename{Nguyen~Hoang \bgroup et al.\egroup }2010]{tuanh2010}
Tu~Anh Nguyen~Hoang, Hoang~Khai Nguyen, and Quang~Vinh Tran.
\newblock 2010.
\newblock An efficient vietnamese text summarization approach based on graph
  model.
\newblock In {\em 2010 IEEE RIVF International Conference on Computing
  Communication Technologies, Research, Innovation, and Vision for the Future
  (RIVF)}, pages 1--6.

\bibitem[\protect\citename{Nguyen-Hoang \bgroup et al.\egroup
  }2012]{nguyen2012tsgvi}
Tu-Anh Nguyen-Hoang, Khai Nguyen, and Quang-Vinh Tran.
\newblock 2012.
\newblock Tsgvi: a graph-based summarization system for vietnamese documents.
\newblock {\em Journal of Ambient Intelligence and Humanized Computing},
  3(4):305--313.

\bibitem[\protect\citename{Ren \bgroup et al.\egroup }2017a]{ren2017}
Pengjie Ren, Zhumin Chen, Zhaochun Ren, Furu Wei, Jun Ma, and Maarten de~Rijke.
\newblock 2017a.
\newblock Leveraging contextual sentence relations for extractive summarization
  using a neural attention model.
\newblock In {\em Proceedings of the 40th International ACM SIGIR Conference on
  Research and Development in Information Retrieval}, SIGIR '17, page 95–104,
  New York, NY, USA. Association for Computing Machinery.

\bibitem[\protect\citename{Ren \bgroup et al.\egroup
  }2017b]{10.1145/3077136.3080792}
Pengjie Ren, Zhumin Chen, Zhaochun Ren, Furu Wei, Jun Ma, and Maarten de~Rijke.
\newblock 2017b.
\newblock Leveraging contextual sentence relations for extractive summarization
  using a neural attention model.
\newblock In {\em Proceedings of the 40th International ACM SIGIR Conference on
  Research and Development in Information Retrieval}, SIGIR '17, page 95–104,
  New York, NY, USA. Association for Computing Machinery.

\bibitem[\protect\citename{Sanh \bgroup et al.\egroup
  }2019]{Sanh2019DistilBERTAD}
Victor Sanh, Lysandre Debut, Julien Chaumond, and Thomas Wolf.
\newblock 2019.
\newblock Distilbert, a distilled version of bert: smaller, faster, cheaper and
  lighter.
\newblock {\em ArXiv}, abs/1910.01108.

\bibitem[\protect\citename{Truong and Nguyen~Quang}2012]{dinh-and-nguyen}
Quoc-Dinh Truong and Dung Nguyen~Quang.
\newblock 2012.
\newblock Một giải pháp tóm tắt văn bản tiếng việt tự động.
\newblock In {\em Hoi thao quoc gia lan thu XV: Mot so van de chon loc cua Cong
  nghe thong tin va truyen thong}, 12.

\bibitem[\protect\citename{Ung \bgroup et al.\egroup }2015]{Ung2015}
Van-Giau Ung, An-Vinh Luong, Nhi-Thao Tran, and Minh-Quoc Nghiem.
\newblock 2015.
\newblock Combination of features for vietnamese news multi-document
  summarization.
\newblock In {\em 2015 Seventh International Conference on Knowledge and
  Systems Engineering (KSE)}, pages 186--191.

\bibitem[\protect\citename{Xu \bgroup et al.\egroup
  }2020]{xu-etal-2020-discourse}
Jiacheng Xu, Zhe Gan, Yu~Cheng, and Jingjing Liu.
\newblock 2020.
\newblock Discourse-aware neural extractive text summarization.
\newblock In {\em Proceedings of the 58th Annual Meeting of the Association for
  Computational Linguistics}, pages 5021--5031, Online, July. Association for
  Computational Linguistics.

\end{thebibliography}

\end{document}